\documentclass[]{elsarticle}

\usepackage{hyperref}

\journal{Image and Vision Computing, SI:ADACV (Elsevier)}

\usepackage{amsmath}
\usepackage{csvsimple}
\usepackage{multirow}
\usepackage{graphicx}
\usepackage{booktabs}
\usepackage{tikz}
\usetikzlibrary{positioning, calc, arrows, shapes, fit}
\usepackage{pgfplots}
\usepgfplotslibrary{colorbrewer}

\usepackage{collcell}
\usepackage{colortbl}

\usepackage{algorithm}
\usepackage{algorithmic}

\usepackage{setspace}

\newcommand*{\MinNumber}{0.0}%
\newcommand*{\MidNumber}{0.3} %
\newcommand*{\MaxNumber}{1.0}%

\newcommand{\ApplyGradient}[1]{%
    \ifdim #1 pt > \MidNumber pt
    \pgfmathsetmacro{\PercentColor}{max(min(100.0*(#1 - \MidNumber)/(\MaxNumber-\MidNumber),100.0),0.00)} %
    \hspace{-0.33em}\colorbox{green!\PercentColor!red}{#1}
    \else
    \pgfmathsetmacro{\PercentColor}{max(min(100.0*(\MidNumber - #1)/(\MidNumber-\MinNumber),100.0),0.00)} %
    \hspace{-0.33em}\colorbox{white!\PercentColor!red}{#1}
    \fi
}
\newcolumntype{G}{>{\collectcell\ApplyGradient}c<{\endcollectcell}}
\renewcommand{\arraystretch}{0}
\usepackage{adjustbox}
\usepackage{array}

\newcolumntype{R}[2]{%
    >{\adjustbox{angle=#1,lap=\width-(#2)}\bgroup}%
    l%
    <{\egroup}%
}
\newcommand*\rotz{\multicolumn{1}{c}}

\begin{document}

\begin{frontmatter}

\title{Improved Multi-Source Domain Adaptation by Preservation of Factors}

\author[TUAddress]{Sebastian Schrom\corref{mycorrespondingauthor}}


\cortext[mycorrespondingauthor]{Corresponding author}
\ead{sebastian.schrom@rmr.tu-darmstadt.de}

\author[HondaAddress]{Stephan Hasler}
\author[TUAddress]{J\"urgen Adamy}

\address[TUAddress]{Control Methods and Robotics Lab, TU Darmstadt, 64283 Darmstadt - Germany}
\address[HondaAddress]{Honda Research Institute Europe, Carl-Legien-Str. 30, 63065 Offenbach - Germany}

\begin{abstract}
        Domain Adaptation (DA) is a highly relevant research topic when it comes to image classification with deep neural networks. Combining multiple source domains in a sophisticated way to optimize a classification model can improve the generalization to a target domain. Here, the difference in data distributions of source and target image datasets plays a major role. In this paper, we describe based on a theory of visual factors how real-world scenes appear in images in general and how recent DA datasets are composed of such. We show that different domains can be described by a set of so called domain factors, whose values are consistent within a domain, but can change across domains. Many DA approaches try to remove all domain factors from the feature representation to be domain invariant. In this paper we show that this can lead to negative transfer since task-informative factors can get lost as well. To address this, we propose Factor-Preserving DA (FP-DA), a method to train a deep adversarial unsupervised DA model, which is able to preserve specific task relevant factors in a multi-domain scenario. We demonstrate on CORe50\,\cite{Lomonaco17}, a dataset with many domains, how such factors can be identified by standard one-to-one transfer experiments between single domains combined with PCA. By applying FP-DA, we show that the highest average and minimum performance can be achieved.
\end{abstract}


\end{frontmatter}

\section{Introduction}

    The optimization of a well generalizing image classifier based on deep learning requires generally a considerable amount of training images that show high variations in visual class appearances. In general each dataset comes with a dataset bias\,\cite{Torralba11} that is caused by a subset of all possible variations. If training data and application data represent samples from the same distribution, i.\,e. have a similar dataset bias, this difference mostly has minor influence on the classification performance. If both are composed of samples from different data distributions we usually speak of domains. In this case a significant decrease in classification performance can be expected if the domains are not handled explicitly\,\cite{Csurka17}.
    The research area of Domain Adaptation (DA) tries to obtain classification models that are able to generalize well to new domains. The new domains are usually called target domains and come with few or no labeled data, while the domains that are used for supervised training of the network are called source domains.

\begin{figure}[tb]
        \centering
        \includegraphics[width=0.975\textwidth]{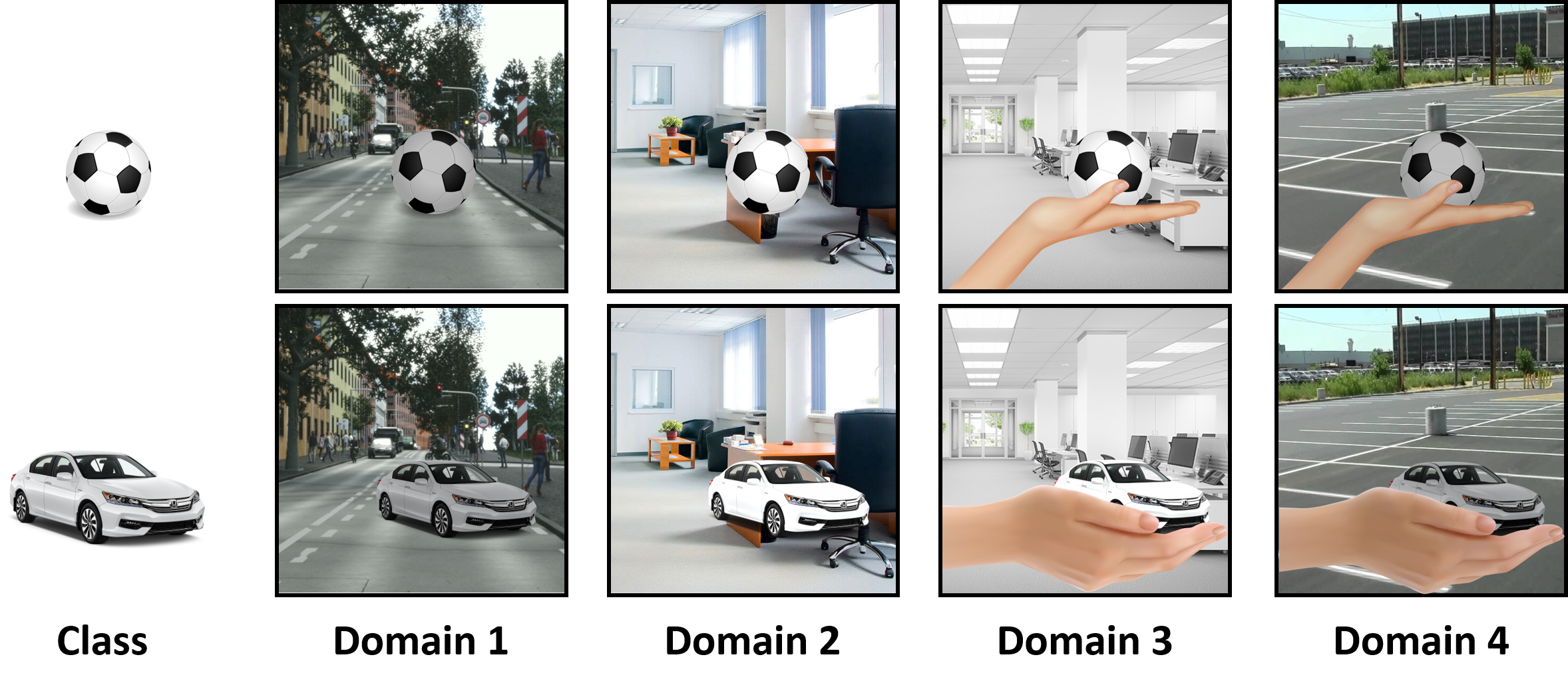}
        \caption{
            The variations in the shown domain dataset can be described by different factors, as e.\,g.\ \textit{'background'}\,$\in$\,\{street, office\,1, office\,2, parking\}, \textit{'location'}\,$\in$\,\{indoor, outdoor\}, \textit{'lighting'}\,$\in$\,\{bright, dark\}, \textit{'hand'}\,$\in$\,\{present, absent\}, \textit{'hand posture'}\,$\in$\,\{flat, hollow\}. We call a factor a domain factor, if it has a constant value within each domain, but different values across all domains. All of the mentioned factors are domain factors, except the \textit{'hand posture'}, whose value changes also along classes.
        }
        \label{fig:Intro_Domainfactors}
    \end{figure}

    The datasets that are used for verification of DA approaches in computer vision, i.\,e the different source and target domains, are often generated under human influence to investigate a specific type of variation that the classification model should deal with. As we will show in this paper, the difference between the datasets is mostly based on values of visual factors, like types of lighting, background or the camera. A domain is consequently defined by a combination of these factors where the value of a factor is consistent in one domain, but can change across domains. We call these factors domain factors. An example is given in Figure\,\ref{fig:Intro_Domainfactors}, where the domain factors are strongly related to recording locations.
    Using such a theory of visual factors allows to describe more formal and understandable aspects of datasets, to predict effects of machine learning, in particular of DA approaches, and to interpret and discuss experimental results with more human interpretability than the unspecific notion of different data distributions allows. 
    
    In general, there are two main types of DA approaches. 
    For the first type, a layer with a simple mathematical operation is designed to normalize out a given aspect explicitly. The parameters of the layer can be determined for the target domain after training. This generally only works for biases that are known or can easily be modeled, and includes methods like dataset based RGB mean subtraction and batch normalization.
    The second type of approaches is based on learned normalization that allows to handle general complex bias differences that can hardly be modeled. Deep learning approaches usually enforce domain invariance at a certain feature layer in the optimization process by an additional loss function. Training usually involves unsupervised data of the target domain, but also generalization of multiple source domains to a completely unknown target domain is expected to improve in this way. In this work we want to focus on such learned normalization approaches.

    Adversarial DA approaches that are based on additionally attached domain classifiers are often used for learned normalization. Their goal is to classify the domain of a sample, however, they are integrated with a specific architectural or loss function modification. With this modification the obtained feature representation should not be informative about the domains given the training data. In other words, what standard adversarial DA tries to do, is to completely remove all domain factors from the feature representation. In this paper we first discuss theoretically that this can be harmful if factors are removed that are generally helpful for classification, and later provide experimental evidence for this so called negative transfer.

    To counteract the negative transfer, we propose Factor-Preserving DA (FP-DA). FP-DA is an adversarial multi-source approach that allows to preserve a given factor by switching off competition between domains that differ in this factor. With this method we report increased classification accuracies in a leave-one-out domain setting, both in terms of average performance, but even more pronounced in the minimal target domain performance, which is the more important measure to look at for safety critical systems.

    The remainder of this paper is structured as follows. We start with a formal definition of domains and its components, followed by the introduction of our theory of visual factors. There we show that scenes captured in images can be described by a superposition of factors and that domains are characterized by domain factors. This is followed by an overview of existing image datasets and DA approaches in the context of visual factors. After this we introduce the frequently used adversarial DA architecture by Ganin et al.\,\cite{Ganin15} on which our adapted training method, FP-DA, is based on. We will prove the advantages of FP-DA using the example of the Core50 dataset\,\cite{Lomonaco17} which comes with 11 domains. Our first experiments investigate how DA influences the transfer in a multi-source domain constellation where 10 source domains and a single target domain are given. Here we will show the effects of negative transfer when naively applying DA. This is followed by single source transfer experiments to identify visual factors that are removed by DA and therefore cause significant errors. We show that Principal Component Analysis (PCA) can be a means for the identification of those. Using the obtained knowledge from these experiments we will finally present the best results with FP-DA in such a multi-source DA scenario.
   
\section{Theory of Visual Factors}
    In this section we will first introduce the formal definition of domains that is mostly used by recent DA approaches. This is followed by the introduction of our theory of visual factors, to which we will refer throughout the remainder of this paper.
    \subsection{Formal Definition of Domains}
    Following the definitions of \cite{Csurka17, Weiss16, Pan10}, we assume to have a set of $n$ data samples $\textbf{X}=\{\textbf{x}_1,\,...,\,\textbf{x}_n\}$, where each sample $\textbf{x}$ represents a multi-dimensional feature vector in the feature space $\mathcal{X}$. Each sample has a corresponding label $y$ from $\mathcal{Y}$ given in $\textbf{Y}=\{\textbf{y}_1,\,...,\,\textbf{y}_n\}$. In classification tasks, the goal is to find the best approximation of the conditional distribution $P(\textbf{Y}|\textbf{X})$ of samples to class labels. However, this approximation strongly depends on the samples included in $\textbf{X}$. DA approaches try to make this approximation more general and evaluate the transfer between sample sets, called domains, that come with different $\textbf{X}$. Specifically here the training and test set are called source and target domain. 
    The general formal definition of source and target domain, as described in \cite{Tommasi13}, is given by $D^s=\{\mathcal{X}^s, P(\textbf{X}^s)\}$ and $D^t=\{\mathcal{X}^t, P(\textbf{X}^t)\}$. Both come with a related task, given by  $\mathcal{T}^s=\{\mathcal{Y}^s, P(\textbf{Y}^s|\textbf{X}^s)\}$ and $\mathcal{T}^t=\{\mathcal{Y}^t, P(\textbf{Y}^t|\textbf{X}^t)\}$, which we assume to be identical here. In this work we consider only homogeneous DA, where the feature spaces $\mathcal{X}^s$ and $\mathcal{X}^t$ are identical, meaning that the difference between domains is related to the marginal distributions, i.\,e.\ $P(\textbf{X}^s)\neq P(\textbf{X}^t)$. This difference is mostly caused by a different sample selection bias\,\cite{Kouw18} for both sample sets and is called covariate shift. A visualization is given in Figure\,\ref{fig:DataDistributions}.
    
    However, the notion of different data distributions does not allow to model the rich constellation of multiple domains comprehensively. Therefore we target to describe domains as combinations of a set of shared underlying factors. Based on relations between factors in the domains the effects of DA approaches can better be predicted and thus be influenced.

    \pgfmathdeclarefunction{gauss}{3}{%
    \pgfmathparse{#3*1/(#2*sqrt(2*pi))*exp(-((x-#1)^2)/(2*#2^2))}
    }
    \tikzstyle {function} = [line width = 2pt]    
    \begin{figure}[tb]
        \centering
        \resizebox{0.35\columnwidth}{!}{
            \begin{tikzpicture}[trim axis right,trim axis left]  
            \begin{axis}[every axis plot post/.append style={
                mark=none,domain=-3:3,samples=50,smooth}, 
            axis x line*=bottom, 
            axis y line*=left, 
            ymax = 0.65,
            ymin = 0,
            xmin = -3,
            xmax = 3,
            xticklabels={},
            ylabel= $P(\tau)$,
            legend pos=outer north east,
            legend cell align={left}
            ] 
            \addplot[function, dashed, blue] {gauss(0,1.1, 1)};
            \addlegendentry{$\tau=\textbf{X}^s$} 
            \addplot[function, red] {gauss(1,0.75, 1)};
            \addlegendentry{$\tau=\textbf{X}^t$}
            \end{axis}
            \end{tikzpicture}
        }
        \caption{Exemplary covariate shift between the marginal distributions of source and target domain sample sets $\textbf{X}^s$ and $\textbf{X}^t$.}
        \label{fig:DataDistributions}
    \end{figure}
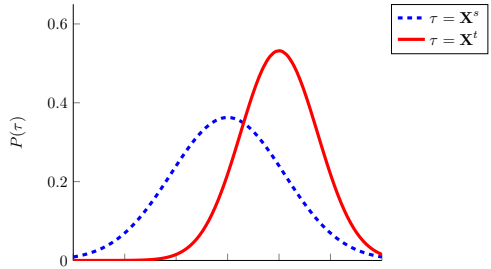

	\subsection{Domains in Context of Visual Factors}
    
    We start from the general notion that each high-dimensional image $\textbf{b}_i$ of a scene $i$ can be generated by a mixing function $M(\textbf{f}_i) = \textbf{b}_i$ from an activation of factors $\textbf{f}_i = (f_{1_i},\, f_{2_i}, \, ..., \, f_{l_i} )$. Each factor $f_{r}$ can be a continuous or categorical variable. In general, many different meaningful combinations of the mixer $M$ and the factors $\textbf{F}$ can be found to model the same image data $\textbf{I}$. Our goal here is not to analytically solve the factorization given a certain optimality criterium. Instead, we like to discuss more theoretically the properties of factors in relation to image classification tasks under domain transfer. This theory we will use later to describe image datasets and predict and influence the effects of DA approaches.

   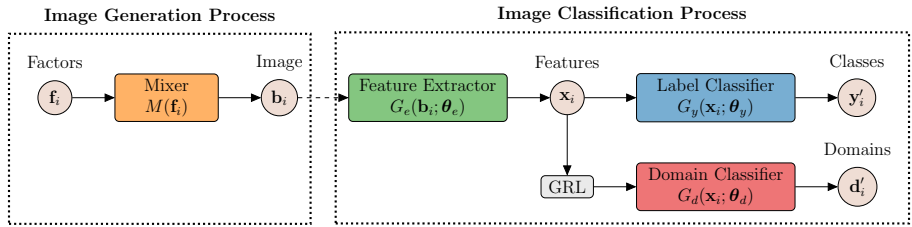
\begin{figure}[tb]
       \centering
       \tikzstyle {arrow_factor} = [-triangle 45,line width = 0.5pt]
\tikzstyle {box_settings} = [minimum width = 2cm, minimum height = 1.0cm, text width=2cm, align= center]
\tikzstyle {circle_settings} = [minimum width = 0.8cm, minimum height = 0.8cm]
\tikzstyle {factor} = [circle, draw, node distance = 1cm, fill=Paired-L, fill opacity = 0.2, text opacity=1, circle_settings]
\tikzstyle {fully} = [fill=Paired-B,fill opacity=0.6, text opacity=1, draw,text width=3.5cm, text centered, rounded corners=.1cm, node distance = 2.3cm]
\tikzstyle {pool} = [fill=Paired-D, fill opacity=0.6, text opacity=1, draw,text width=2.2cm,  text centered, rounded corners=.1cm]
\resizebox{\columnwidth}{!}{
    \begin{tikzpicture}[font=\large]
    
    \node[](F1) at (0,0) {};
    \node[factor, below = of F1, yshift= 0.5cm] (F2) {$\textbf{f}_i$};
    
    \node[pool, fill=Paired-H, rectangle, draw, right = of F2, node distance = 1cm](Mixer) {Mixer $M(\textbf{f}_i)$};
    
    \node[circle, draw, fill=Paired-L, fill opacity = 0.2, text opacity=1, right = of Mixer, circle_settings](Image) {$\textbf{b}_i$};
    \node[above = of Image, anchor = north, yshift=-0.3cm]{Image};
    
    \node[pool, right of= Image, text width=3.5cm, node distance = 3.5cm](FeatExt) {Feature Extractor \\ $G_e(\textbf{b}_i;\boldsymbol{\theta}_e)$};
    
    \node[circle, draw, fill=Paired-L, fill opacity = 0.2, text opacity=1, right = of FeatExt, circle_settings, node distance=3.5cm](FeatVect) {$\textbf{x}_i$};
    \node[above = of FeatVect,   anchor = north, yshift=-0.3cm]{Features};
    
    \node[fully, right of= FeatVect, node distance = 3.5cm](Classifier) {Label Classifier \\ $G_y(\textbf{x}_i;\boldsymbol{\theta}_y)$ };
    
    \node[fully, text width=1cm, draw, fill=gray!80, fill opacity = 0.2, text opacity=1, below of=FeatVect, xshift = 0.0cm, node distance = 2.1cm](circle){GRL};
    
    \node[fully, fill = Paired-F, below of= Classifier, node distance = 2.1cm](D_Classifier){Domain Classifier \\ $G_d(\textbf{x}_i;\boldsymbol{\theta}_d)$ };

    \node[circle, draw, fill=Paired-L, fill opacity = 0.2, text opacity=1, right = of Classifier, circle_settings](Class) {$\textbf{y}'_i$};
    \node[above = of Class,   anchor = north, yshift=-0.3cm]{Classes};

    \node[circle, draw, fill=Paired-L, fill opacity = 0.2, text opacity=1, right = of D_Classifier, circle_settings](D_Class) {$\textbf{d}'_i$};
    \node[above = of D_Class,   anchor = north, yshift=-0.3cm](Domains_title){Domains};
    
    \node[above = of F2, anchor = north,  yshift = -0.3cm](Factors_title){Factors};

    \node[circle, left of = circle, circle_settings, xshift = -7cm](helper1) {\color{white}$\textbf{d}_i$$'$};
    \node[right of  = Factors_title, xshift = 7cm](helper2){\color{white}Factors $\textbf{f}_i$};
    
    \node[line width=0.5mm, draw, dotted, fit=(Factors_title) (F2) (Image) (helper1), inner xsep = 0.3cm, inner ysep = 0.4cm](ImageProcess){};
    \node[above = of ImageProcess, anchor = north,  yshift = -0.3cm]{\textbf{Image Generation Process}};
    
    \node[line width=0.5mm, draw, dotted, fit=(FeatExt) (Domains_title) (D_Class) (helper2), inner xsep = 0.3cm, inner ysep = 0.4cm](ImageProcess2){};
    \node[above = of ImageProcess2, anchor = north,  yshift = -0.3cm]{\textbf{Image Classification Process}};
    
    \draw [arrow_factor] (F2)  -- (Mixer.west);
    \draw [arrow_factor] (Mixer.east)  -- (Image);
    \draw [arrow_factor, dashed] (Image.east)  -- (FeatExt);
    \draw [arrow_factor] (FeatExt.east)  -- (FeatVect);
    \draw [arrow_factor] (FeatVect.east)  -- (Classifier);
    \draw [arrow_factor] (Classifier.east)  -- (Class);
    \draw [arrow_factor] (FeatVect)  -- (circle);
    \draw [arrow_factor] (circle)  -- (D_Classifier);
    \draw [arrow_factor] (D_Classifier)  -- (D_Class);
    
    \node[](origin) at (0, 0){};
    \node[](right) at (6, 0){};
    
    \end{tikzpicture}
}
       \caption{An image $\textbf{b}_i$ can be generated by a function M from an activation of factors $\textbf{f}_i$. For image classification, a feature vector $\textbf{x}_i$ can be extracted from $\textbf{b}_i$, and $\textbf{y}_i$$'$ can be predicted by a label classifier. Adversarial DA approaches often use an additional domain classifier that is connected via a gradient reversal layer (GRL).}
       \label{fig:Image_Factors}
   \end{figure}
    %
    For the classification of an image $\textbf{b}_i$, a feature extractor usually transforms the image to a feature vector $\textbf{x}_{i}$, ideally of a feature space where classes can easily be separated. 
    Based on $\textbf{x}_i$, a label classifier predicts the class probabilities $\textbf{y}'_i$.
    In adversarial DA approaches often an additional domain classifier is attached to the feature extractor via a gradient reversal layer, which influences only backpropagation.
    The overall process with image generation and classification is depicted in Figure\,\ref{fig:Image_Factors}.

    In general, the factors can be categorized as task-informative or not task-informative. Task informative factors strongly correlate with a certain class, but are not necessarily directly related to that class. Such as for example the factor \textit{'hand posture'}, which can give clues about the ground-truth class of a presented object.
    
    For domain classification, some factors can further be categorized as domain-informative. Domain-informative factors have independent of the ground-truth class a constant value for images from a certain domain, which changes across domains. Such factors are often associated to fixed recording locations of each domain, which consequently induce changing lighting and background related factor values across domains. A domain itself therefore can be described by a combination of factors whose values are constant within each domain, but can differ across domains. We call these factors the domain factors.
    A domain invariant feature extractor must therefore not consider domain factors, since these can lead to misclassification if a classifier uses features based on them and the domain changes. However, removing domain factors from the feature representation can also have adverse effects as we will show.

    In general there are infinite possibilities on how to factorize scenes by visual factors.
    The factorization could for example either be human designed, by choosing certain factors manually, or estimated by a component analysis method based on a given image dataset. The choice ultimately influences the number of factors and the nameability, which is not necessarily given for the latter. Other aspects like the mutual dependence of individual factors, which consequently influences their values, and the controllability of a factor in a scene are also dependent on the chosen factorization. In the end, a well-chosen factorization might provide increased predictability and explainability of classification models handling new factor values or new combinations of known values. However, this also depends on the fineness of the factorization. While for some a discrimination of indoor and outdoor is enough, for others a more fine-grained factorization is required.

    Exemplary views on how to factorize scenes can be found in scene rendering or photography.
    In the rendering view the factorization is based on human interpretable factors:
        \begin{itemize}
            \item Object factors: \textit{'type'}, \textit{'shape'}, \textit{'color'}, \textit{'texture'}, \textit{'viewpoint'},\,... 
            \item Light source factors: \textit{'type'}, \textit{'positioning'}, \textit{'emission power'},\,... 
            \item Camera factors: \textit{'sensor'}, \textit{'filter'}, \textit{'processing'}, \textit{'viewpoint'},\,...
        \end{itemize}
    The factorization and with it the number of factors is pre-defined by the rendering program. The factor values and their mutual dependence are fully controllable by a rendering artist. He decides for instance how chosen object colors appear in combination with placed light sources in the rendered scene. 
           
    The photographer view differs from the rendering view in reduced controllability and increased mutual dependence of the factors. While in natural images, the object in focus and the viewing angle are usually controlled by the photographer, there is less control over natural lighting conditions and context objects. Certain objects usually induce other context objects, just as a car would most likely induce asphalt.
    
    The two presented views describe extremes regarding controllability of factor values. Sometimes, image datasets, especially those used for DA, describe a mixture of these. This is for example the case when objects are artificially placed in unnatural environments, where the target object related factors are fully controlled but the environment factors are not.

    \subsection{Effects of Machine Learning and DA in Context of Visual Factors}

\begin{figure}[tb]
        \centering
        \includegraphics[width=0.94\textwidth]{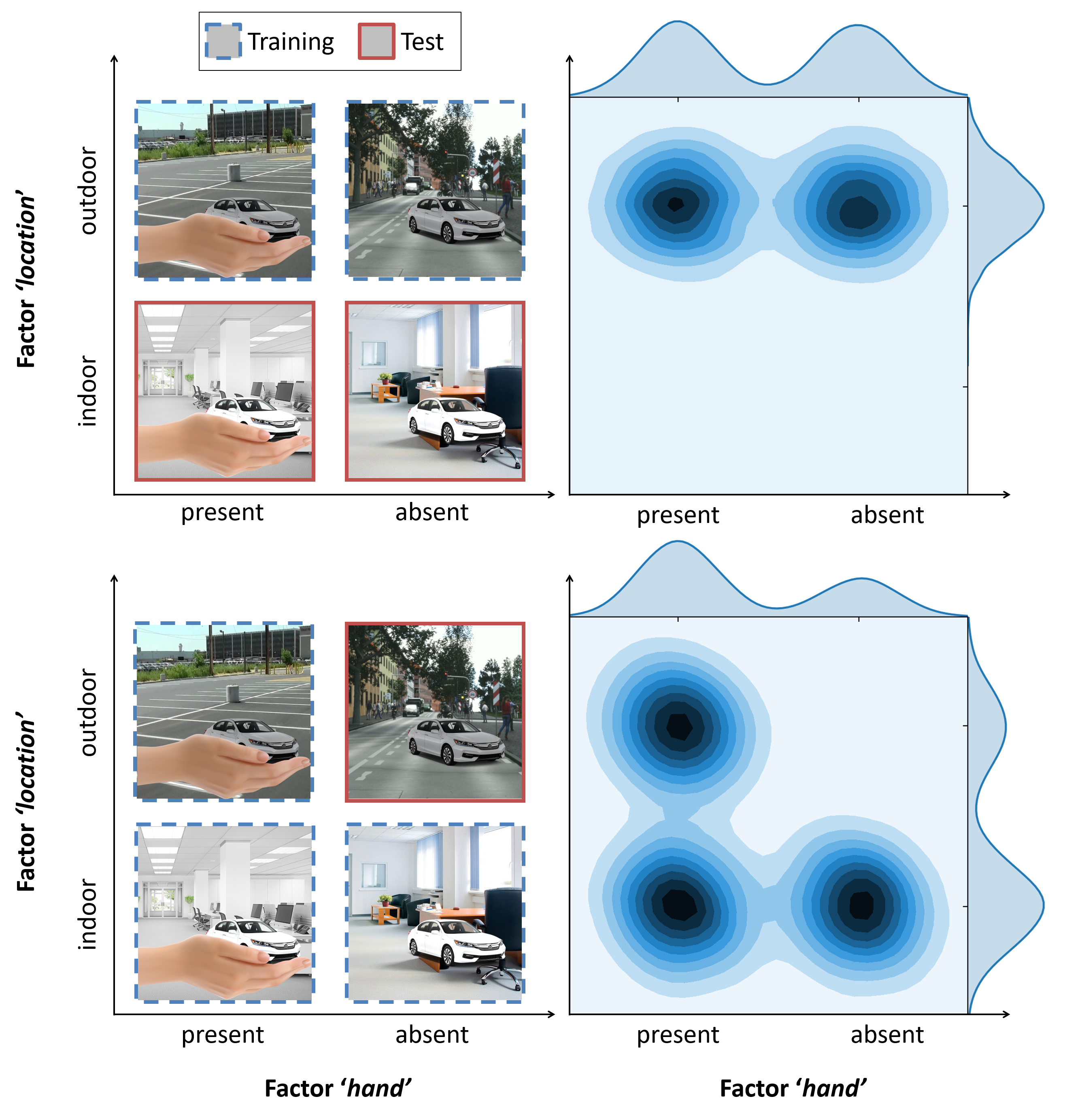}
        \caption{Constellations of factor values in training and test set. \textit{Top:} New value for factor \textit{'location'} in the test set. \textit{Bottom}: New combination of factor values in the test set. All factor values are already included in the training set.}
        \label{fig:Factors_both}
    \end{figure}

    In general, a test set defines what changes regarding factor values a classification model should be able to generalize to. First, there may be new, unseen factor values for certain class samples in the test set, and then there could be new combinations of already seen factor values. Both cases are depicted in Figure\,\ref{fig:Factors_both}.

    New, unseen values in the test set could be related to factors of target objects, like different object colors, as well as to context objects factors, lighting factors, or others. A classification model requires internal class representations that ignore such factors to perform well on the test set.
    However, standard machine learning algorithms that are optimized on a training set are not forced to be invariant of such factors and could therefore misclassify samples if these factors are considered for classification. In DA, the information about the new factor values is often implicitly provided by unsupervised samples from the target domain. Using these samples in a sophisticated way during optimization can help to become invariant of the changing factors.

    A second case is the occurrence of new combinations of values, while each value was already seen in other combinations during training.
    With standard machine learning it is likely that combinations are learned as a single representative feature, if factor values are exclusively shown in combination with specific values of other factors. Consequently, the learned model might fail on new combinations. If the factors on which the combinations are based are domain factors, then they will be removed from the feature representation by DA already during training. Thus, there will be no influence of the new combinations in the test set.

    A general negative side effect of DA removing domain factors occurs when the removed factors are not only domain-informative, but also task-informative or visually close to other task-informative factors. For example, if the hand, which holds the target object, is removed from the feature representation, the factor \textit{'hand posture'} could also be removed, which, as mentioned before, can be task-informative as well. Additionally, in case of visual closeness this might remove parts of the target object due to network architectural reasons related to receptive field sizes.

\section{Related Work}
    In the first part of this section we will analyze different image datasets in relation to the concept of the introduced visual factors. The second part will review different related single and multi-source DA approaches.
    
\textbf{Image Datasets in Context of Visual Factors.}
    There are many datasets for image categorization, that are based on internet search, like \cite{imagenet09, CaltechDataset07}. This provides a large variation for the different categories in terms of objects and their context. However, these datasets also come with strong biases, like a western world bias\,\cite{DeVries19, shankar17} or a capture bias\,\cite{Torralba11}, and do not allow control over many factor values, like the object type or its viewing point.

    In contrast, datasets used for DA research \cite{Peng15, Li17, Koniusz18, Peng18} usually come with increased control over specific factor values. 
    There are datasets like \cite{Venkateswara17, Saenko10, Bergamo10, Peng19} where some domains are based on restricted internet searches, as for example only images from Amazon or clipart type. However, the control over object instances is still limited.
    Increased control over objects and background is usually found in the robotics community, where the same objects are placed in front of different backgrounds, like in CORe50\,\cite{Lomonaco17}, but also in the well-known Office-31 DA dataset\,\cite{Saenko10} where two of the three domains were generated similarly. If the segmentation of the object is known, like in MNIST\,\cite{Lecun98}, the background can virtually be changed as in MNIST-M\,\cite{Ganin15} to obtain a new domain.   
    Full control over all factor values is usually found in rendered datasets, like the disentanglement library\,\cite{Locatello19}, which shows simple objects in a 3D environment with changing values for factors like size, viewing angle, color and others.

    \textbf{Recent Approaches in Context of Visual Factors.}  
    Recent works in DA are mostly based on either kernel-based methods or adversarial methods. Kernel-based methods, like maximum mean discrepancy\,\cite{Gretton12}, try to align feature distributions of the involved domains by using a cost function on a shared feature extractor that should minimize the distance between distributions. This is applied for example in \cite{Tzeng14, Long15, Long16, Long19, Bousmalis16, Karimpour20}. Approaches using adversarial methods\,\cite{Tzeng15, Ajakan14} are mostly based on a regressor or a full domain classifier that is additionally attached to the shared feature extractor. The overall target here is that the feature activations of the shared feature extractor should not rely on domain-informative factors. This is often achieved by flipping the gradient from the domain classifier at the connection to the shared feature extractor via a gradient reversal layer\,\cite{Ganin15, Ganin16, Bousmalis16, Long18}. Other approaches, like \cite{Tzeng17}, optimize a first feature extractor supervisedly on the source data and after this train a second feature extractor on the unsupervised target data, in a way that a classifier cannot determine the domain from their outputs. This is similar to the workflow of generative adversarial networks\,\cite{Goodfellow14}.
    
    Newly proposed approaches are usually only tested on data for which a gain in generalization performance can be reported. The effect of negative transfer\,\cite{Rosenstein05, Weiss16} is only rarely addressed, as e.\,g.\ in \cite{Schrom17, Schrom19, Cao19, Ge14}. Research showed that it can occur when the label spaces of source and target domain differ \cite{Cao19}, or the distributions are imbalanced \cite{Ge14}. In \cite{Pei18} they state that there exist underlying multimode structures of the data distributions of domains that can lead to negative transfer and tackle this by using individual domain classifiers for each class.

    We see a general problem in the removal of all domain factors, which tends to remove task-informative factors as well. Having multiple source domains usually leads to the removal of more domain factors and therefore a pronounced negative effect can be expected. The negative effect of irrelevant source domains was reported in \cite{Ge14, Duan12}. Our FP-DA is designed to reduce negative transfer by switching off competition between groups of domains. There are other methods that try to limit or better control the competition between domains.
    This is mostly done by a weighting strategy that is a based on a score that rates the similarity of domains, either pair-wise, i.e. only competition between the target domain and a single source domain, as in \cite{Guo18, Mansour08, Redko19, Zhao20, Xu18} or as a weighted competition also between multiple source domains, as in \cite{Li18, Rakshit19, Peng19, Wang19}. However, none of the mentioned approaches consider factor or group based competitions as we do.

\section{Our Approach} \label{sec:ourApproach}
    The main idea of our FP-DA is to preserve a chosen factor $f_{c}$ during DA. 
    It is based on the adversarial DA architecture from \cite{Ganin15}. The architecture corresponds to the typical adversarial one shown in Figure \ref{fig:Image_Factors}, where a domain classifier is used together with a gradient reversal layer to inhibit learning of domain-informative features in the feature extractor. The parameters of the feature extractor $\boldsymbol{\theta}_{e}$, the label classifier $\boldsymbol{\theta}_{y}$, and of the domain classifier $\boldsymbol{\theta}_{d}$ are updated in the following way:
    \begin{align}
    \boldsymbol{\theta}_{e}\quad &\leftarrow \quad \boldsymbol{\theta}_{e} - \mu ( \frac{\partial L_{y}}{\partial \boldsymbol{\theta}_{e}} - \lambda \frac{\partial L_{d}}{\partial \boldsymbol{\theta}_{e}}) \label{eq:featureExtractorParameter}\\
    \boldsymbol{\theta}_{y}\quad &\leftarrow \quad  \boldsymbol{\theta}_{y} - \mu \frac{\partial L_{y}}{\partial \boldsymbol{\theta}_{y}}\\
    \boldsymbol{\theta}_{d}\quad &\leftarrow \quad \boldsymbol{\theta}_{d} - \mu \frac{\partial L_{d}}{\partial \boldsymbol{\theta}_{d}}      \label{eq:domainClassifierParameter}
    \end{align}
    with $L$ being the corresponding loss for each branch of the architecture and $\mu$ the learning-rate. The influence of the domain classifier on the feature extraction path is controlled by the adaptation factor $\lambda$ which is smoothly increased during optimization. For further details on the general training procedure we refer to \cite{Ganin15}. 
    Originally the domain classifier was used to discriminate data from a single source and target domain, however, it can also be used for multiple source domains and in this case also without any data from the target domain. 
    In each case, all domains compete against each other in the domain classifier.
    
\begin{figure}[tb]
    \centering
    \includegraphics[width=\textwidth]{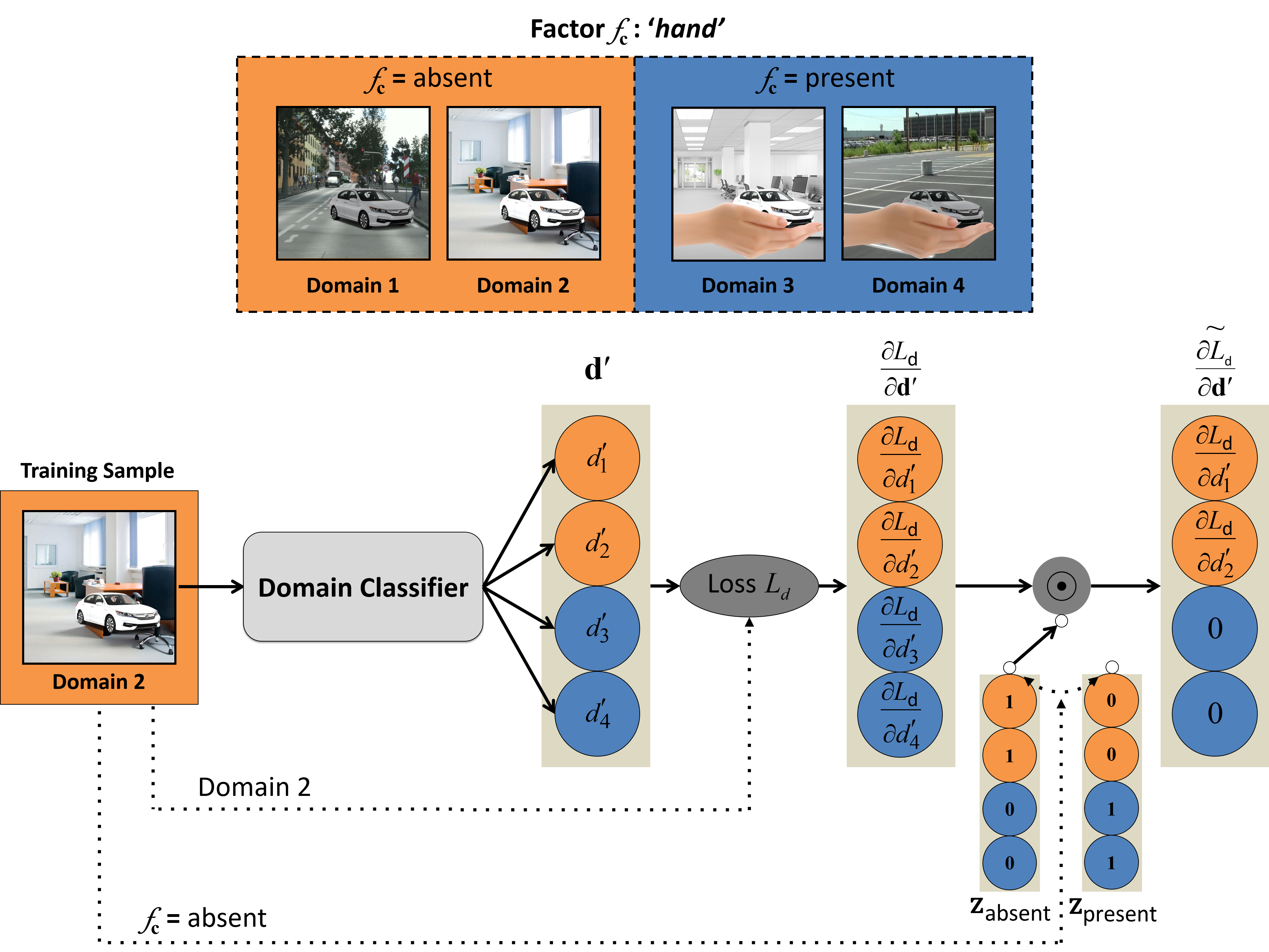}
    \caption{Our FP-DA training method uses given groups of domains, here generated by the values of the chosen factor $f_c$, \textit{'hand'}, and allows competition only between domains within each group. A training sample from the group where $f_c$=\textsf{absent} is forwarded through the domain classifier. The resulting gradient vector $\frac{\partial L_{d}}{\partial \textbf{d}'}$ is element-wise multiplied by $\textbf{z}_{\textsf{absent}}$ that zeros the gradients of the domains where $f_c$=\textsf{present}.
    }
    \label{fig:FPDA}
    \end{figure}

    To preserve a given factor, in FP-DA we switch off competition between domains that have different values for a chosen factor $f_{c}$. For this we use the decomposition of the gradient
    \begin{align}
    \frac{\partial L_{d}}{\partial \boldsymbol{\theta}_{d}} = \frac{\partial L_{d}}{\partial \textbf{d}'} \cdot \frac{\partial \textbf{d}'}{\partial \boldsymbol{\theta}_{d}}, 
    \end{align}
    and replace $\frac{\partial L_{d}}{\partial \textbf{d}'}$ by 
    \begin{align}
    \newcommand\Wtilde{\stackrel{\widetilde{\hphantom{aa}}}{\smash{\frac{\partial L_{d}}{\partial \textbf{d}'}}\rule{0pt}{2.6ex}}}
    \Wtilde = \textbf{z} \odot \Big(\frac{\partial L_{d}}{\partial \textbf{d}'}\Big), 
    \end{align}
    where $\textbf{z}$ is an $m$-dimensional vector corresponding to the $m$ domains. If an element $z_j=1$ then the gradient of the domain neuron $d_j$ is kept and propagated to the parameters $\boldsymbol{\theta}_{d}$. If $z_j=0$ then no gradient is propagated from the neuron $d_j$.
    For a given training sample of domain $k$, in FP-DA we compute $z_j$ like 
    \begin{equation}
        z_j = 
        \begin{cases}
        1 & \text{if } f_{c}(j) = f_{c}(k) \\
        0 & \text{otherwise,}
        \end{cases}
    \end{equation}
    to only allow competition between domains that have the same value for $f_c$ as the current training sample. Note, for FP-DA we assume $f_c$ to be a categorical variable. The principle of FP-DA is exemplarily depicted in Figure\,\ref{fig:FPDA}.
    A limitation of FP-DA is that within each group of domains that share the same value for $f_c$, all other domain factors from the dataset must still be represented to learn a general model.
    \section{Experiments}
    In this section we will evaluate FP-DA at the example of an object classification task in a multi-source domain scenario. For this we will first naively apply the standard DA method in leave-one-out experiments, where we choose one domain as target domain and the rest as source domains, to investigate the effects of removing all domain factors. This is followed by one-to-one experiments, i.\,e.  single source transfer, where we investigate the transfer performance in context of visual factors. Based on this we will show how factors responsible for high transfer errors can be identified and used for the grouping that is required in FP-DA. Finally we will show how our method is able to improve the average and minimum performance in such a multi-source domain scenario. 
    Note, besides the leave-one-out and one-to-one experiments further constellations in between are possible, however, we decided for the mentioned, since those are the most meaningful cornercases and also tractable to do.
    \subsection{Experimental Setting}\label{sec:Core50_Intro}
\begin{figure}[tb]
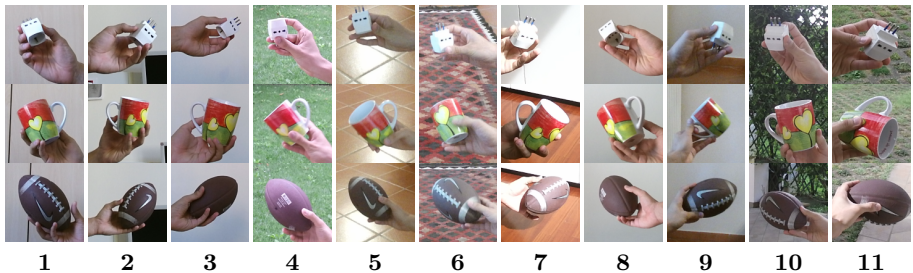

    \centering
            \foreach \y in {{1, 2, 3, 4, 5, 6, 7, 8, 9, 10, 11}}{
                \foreach \x in \y {
                    \begin{tikzpicture}
                    \node (j) [inner sep=0pt] {\includegraphics[width=.085\linewidth] 
                        {Images/Plug_allSessions/S\x_plug.png}};
                    \node (k) [inner sep=0pt, below of = j, yshift=-0.05cm] {\includegraphics[width=.085\linewidth] 
                        {Images/Mug_allSessions/S\x_mug.png}};
                    \node (l) [inner sep=0pt, below of = k, yshift=-0.05cm] {\includegraphics[width=.085\linewidth] 
                        {Images/Football_allSessions/S\x_ball.png}};
                    \node [below of = l, node distance = 0.8cm] {\small{\textbf{{\x}}}};
                    \end{tikzpicture}
                    \hspace{-0.25cm}
                }\\
            }
    \caption{Exemplary images from the 11 different sessions of the CORe50 dataset. The background of each session is representative for each of the 50 objects. In each session a single hand, i.\,e. right or left, was used exclusively for presentation. We use each session as an individual domain.}
    \label{fig:OverviewSessions}
    \end{figure}

    \textbf{Dataset.}
    For our experiments we chose the CORe50 dataset introduced in \cite{Lomonaco17}. The dataset consists of 11 distinct recording sessions, where 50 objects, divided into 10 categories with 5 objects each, were captured while held and rotated in a human hand. From each session a representative image of three objects is shown in Figure \ref{fig:OverviewSessions}. 

    Originally this dataset was meant to serve as a benchmark for continuous learning, where class instances from new sessions were added over time and the task was to classify the instance or the category. Nevertheless, we see this dataset with its controlled and clean recording setting also applicable for domain transfer investigations. Therefore we will use each session as a domain and choose category prediction as our main classification task for all experiments. To our knowledge we are the first to use it in the domain transfer context. It can be described mainly by the photographer view since it is based on natural images, however, not fully, since the objects were artificially placed in the environment. The domain factors here are mostly related to the recording location. Since the objects are in each domain the same, they can be described as a constant factor across domains. Clearly each domain was exclusively captured holding the object with either the right or the left hand. However, it remains unclear whether always the same person was presenting the object, if not, this could describe another domain factor that is related to individual styles of presenting and holding the objects.

    Since there is only a single recording stream per domain, no clear training/test separation is possible. This would be in general no problem since most of our results focus on generalization to a completely unknown target domain. However, there are some results where unsupervised data of the target domain is used during training. Therefore we chose from each session the half of the 300 images per object as fixed test data, by dividing the recording stream into chunks of 20 images. Note, when training on domain A and testing on domain B, independent of the application of DA, we use all images of A, but only 50\% of B, since the other 50\% are reserved as unsupervised data for DA.

    \textbf{Implementation Details.}
    For the baseline experiments without DA we used the standard VGG-16 architecture from \cite{Simonyan14}. The weights are initialized by a publicly available set of parameters that was pre-trained on the ImageNet database.
    For DA experiments we attached additionally the domain classifier after the last convolutional layer of this architecture. It consists of two fully connected layers with 1024 neurons each, similar to \cite{Ganin15}, and a final output layer with $m$ neurons, corresponding to the number of domains. 
    The adaptation factor $\lambda$ is smoothly increased during optimization from 0 to 1 with the same update rule as presented in \cite{Ganin15}.  
    We chose a batch-size of 64 and an initial learning-rate of $\mu=0.0001$, which was decreased over the number of steps. On both classifiers we used dropout in the fully-connected layers and applied softmax on the final output. The input samples are mean normalized, but not modified by any data augmentation method. The architecture was implemented in TensorFlow.
    For all reported numbers we trained the specific architecture 10 times with randomly initialized parameters $\boldsymbol{\theta}_d$ of the domain classifier and randomly shuffled training data, for two epochs and averaged the classification accuracy. 
    \subsection{Standard Multi-Source Transfer}

    For the given dataset we first do leave-one-out experiments to find out how well many source domains can generalize to a single target domain. This we do without and with DA. For the latter case we only use the source domain data or we further provide unsupervised data of the target domain.
    Even without DA a good performance can be expected, as most domains are similar to at least one other domain. DA without unsupervised target data should then improve generalization for unseen combinations of factor values, while using target data should further help in the case of new factor values.
    \begin{figure}[tb]
       	\centering
       	\def\plotheight{0.5\columnwidth} 
\def\plotwidth{1\columnwidth} 
\def\legendscale{0.7}
\def\plotTextSize{}
\def\drawParam{none}
\pgfplotsset{every tick label/.append style={font=\footnotesize}} 
\begin{tikzpicture}[font=\plotTextSize]
		\begin{axis}[
        ybar=0pt,
        bar width=2.8pt,
		legend style={nodes={scale=\legendscale, transform shape}},
		height=\plotheight,
		width=\plotwidth,
		grid=minor,
		ymin = 70,
		ymax = 100,
		xtick={1,2, 3, 4, 5, 6, 7, 8, 9, 10, 11, 12, 13, 14},
        xticklabels={1, 2, 3, 4, 5, 6, 7, 8, 9, 10, 11, , \footnotesize \textit{Avg}, \footnotesize \textit{Min}},
        minor xtick={1.5, 2.5, 3.5, 4.5, 5.5, 6.5, 7.5, 8.5, 9.5, 10.5, 11.5, 12.5, 13.5},
        minor ytick={80, 90},
        major tick length=0pt,
        xtick pos =left,
        ytick pos = left,
		xlabel style={text width=9cm, align=center},
		xlabel= Target Domain,
		ylabel= Classification Accuracy,
   		ylabel style = {yshift=-0.4cm},
		legend pos=south west,
		legend cell align={left},
        legend columns=3,
        legend style={/tikz/every even column/.append style={column sep=0.5cm}},
        legend style={at={(0.5,1.025)},anchor=south},
        enlarge x limits = {abs=15pt},
		yticklabel=\pgfmathparse{\tick/100}\pgfmathprintnumber{\pgfmathresult}\,,
		]
		\centering
		\addplot[fill=Paired-B, draw = \drawParam] coordinates {
            (1, 94.32)
            (2, 91.30)
            (3, 94.9)
            (4, 93.0)
            (5, 90.6)
			(6, 75.3)
			(7, 89.02)
			(8, 94.40)
			(9, 90.8)
            (10, 81.1)
            (11, 92.9)
            (13, 89.78) 
            (14, 75.3) 
		};
		\addlegendentry{no DA}

        \addplot[fill=black, draw = \drawParam] coordinates {
            (1, 92.6)
            (2, 91.8)
            (3, 92.9)
            (4, 92.3)
            (5, 91.8)
            (6, 76.2)
            (7, 87.54)
            (8, 94.0)
            (9, 89.7)
            (10, 80.2)
            (11, 91.9)
            (13, 89.17) 
            (14, 76.2)
        };
        \addlegendentry{DA\,\cite{Ganin15} (w/o usv)}

        \addplot[fill=Paired-C, draw = \drawParam] coordinates {
            (1, 92.1)
            (2, 92.9)
            (3, 92.6)
            (4, 90.5)
            (5, 91.9)
            (6, 86.34)
            (7, 87.27)
            (8, 92.39)
            (9, 85.7)
            (10, 87.03)
            (11, 89.91)
            (13, 89.87) 
            (14, 85.7)
        };
        \addlegendentry{DA\,\cite{Ganin15} (w/ usv)}

		\end{axis}
\end{tikzpicture}
       	\caption{Multi-source leave-one-out experiments. The left out domain represents the target domain, while the others are used as the supervised source domains. The experiments were carried out with and without standard DA, while the latter was applied with and without additional unsupervised (usv) samples from the target domain.}
       	\label{fig:leave_one_out_no_specialDA}
    \end{figure}
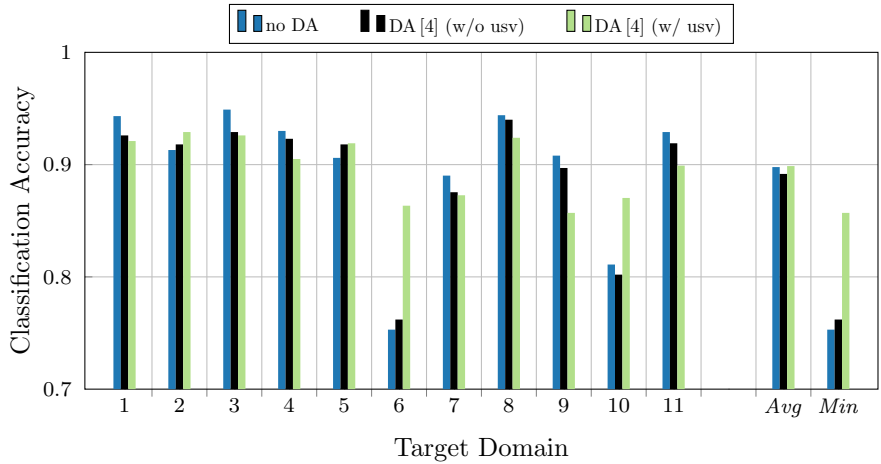

    Looking at the results without DA in Figure \ref{fig:leave_one_out_no_specialDA}, we see that the expectations are widely met. The classification accuracy on the target domains reaches mostly more than 90\,\%. The only target domains that show a comparatively poor performance are domains 6 and 10. Looking at their visual characteristics, we can see that they differ significantly from the others by having both strong fine-grained edges in the background. In domain 6 caused by a carpet and in 10 caused by a fence. Interpreting these aspects as factor values, it is unclear whether these are new values or new combinations of known values.
    
    The results for DA without unsupervised data show for most of the target domains a slight drop in performance indicating negative transfer. This is can be on the one hand site due to the additional constraint on the feature extractor by the gradient from the domain classifier, but also due to task-informative factors that are removed now. On domain 6 and 10 the accuracy has also only slightly changed, which suggests that the issue here are mainly completely new factor values and not combinations.
    
    Looking at the results of DA with unsupervised data, the assumption about new factor values in domain 6 and 10 can be approved with a significant increase in accuracy.
    However, for other experiments, like for example on domain 1, 4, and 8,  we mostly see a weaker performance compared to DA without unsupervised data. We believe that because of the increased number of domains during training, even more factors, including task-informative factors, are now removed from the feature representation.
    Compared to experiments without DA, there is a clear performance drop on seven target domains. We explain this also by the additional constraint and removed task-informative factors.
    
    When comparing the results without DA to DA with unsupervised data in total, the negative transfer for some domains balances out the strong gain for domains 6 and 10 on average, while the latter leads to a strongly improved minimal performance. We think that the negative transfer can be reduced by preserving certain factors. These factors are based on mutual similarity and difference between individual domains and thus can not easily be inferred from the multi-source setting here.

\subsection{Single-Source Transfer without Domain Adaptation}
    To determine error causing factors in DA we conducted one-to-one transfer experiments without DA first, i.\,e.\ optimized the VGG-16 architecture on each individual domain and evaluated it on all others.
    The results are given in Table\,\ref{tab:noDA}. 
    For each target domain we also give the maximal performance among all source domains. These maximal values are all smaller than their multi-source counterpart in Figure\,\ref{fig:leave_one_out_no_specialDA}. This means that multi-source training really builds a more general model instead of just memorizing all domains individually.

        \def\avgColor{gray!30}
    \def\hdColor{Paired-B!60}
    \setlength{\tabcolsep}{0.2pt} 
    \renewcommand{\arraystretch}{1.1} 
    \begin{table}[tb]
        \centering
        \resizebox{\columnwidth}{!}{
        \begin{tabular}{cc*{11}{G}c@{\hskip 0.05in}c}
            \multicolumn{2}{c}{} & \multicolumn{11}{c}{\textbf{Target Domain}} \vspace{0.1cm} \\ 
            &\multicolumn{1}{c}{}& \rotz{\textbf{1}} &
             \rotz{\textbf{\colorbox{\hdColor}{2}}} &
             \rotz{\textbf{\colorbox{\hdColor}{3}}} &
             \rotz{\textbf{4}} &
             \rotz{\textbf{5}} &
             \rotz{\textbf{6}} &
             \rotz{\textbf{\colorbox{\hdColor}{7}}} &
             \rotz{\textbf{8}} &
             \rotz{\textbf{\colorbox{\hdColor}{9}}} &
             \rotz{\textbf{10}}& 
             \rotz{\textbf{\colorbox{\hdColor}{11}}}&
             \hphantom{a}\textit{Avg}\hphantom{a} &
             \textit{Min}\\ \cmidrule{3-13}
            \parbox[t]{4mm}{\multirow{11}{*}{\rotatebox[origin=tr]{90}{\textbf{Source Domain}}}}
            &\textbf{1}&\rotz{-}&0.67&0.72&0.74&0.68&0.45&0.64&0.80&0.57&0.48&0.52&\multicolumn{1}{c}{\colorbox{\avgColor}{0.63}}&0.45\\
            &\textbf{\colorbox{\hdColor}{2}}&0.69&\rotz{-}&0.80&0.59&0.54&0.55&0.72&0.60&0.76&0.52&0.69&\multicolumn{1}{c}{\colorbox{\avgColor}{0.65}}&0.54\\
            &\textbf{\colorbox{\hdColor}{3}}&0.69&0.78&\rotz{-}&0.52&0.51&0.36&0.72&0.60&0.74&0.40&0.66&\multicolumn{1}{c}{\colorbox{\avgColor}{0.60}}&0.36\\
            &\textbf{4}&0.70&0.46&0.50&\rotz{-}&0.68&0.70&0.51&0.72&0.36&0.68&0.52&\multicolumn{1}{c}{\colorbox{\avgColor}{0.58}}&0.36\\
            &\textbf{5}&0.67&0.42&0.46&0.68&\rotz{-}&0.62&0.40&0.72&0.37&0.54&0.34&\multicolumn{1}{c}{\colorbox{\avgColor}{0.52}}&0.34\\
            &\textbf{6}&0.68&0.46&0.44&0.72&0.68&\rotz{-}&0.47&0.67&0.38&0.54&0.49&\multicolumn{1}{c}{\colorbox{\avgColor}{0.55}}&0.38\\
            &\textbf{\colorbox{\hdColor}{7}}&0.70&0.75&0.75&0.58&0.52&0.47&\rotz{-}&0.65&0.66&0.47&0.76&\multicolumn{1}{c}{\colorbox{\avgColor}{0.63}}&0.47\\
            &\textbf{8}&0.83&0.59&0.57&0.76&0.71&0.59&0.57&\rotz{-}&0.48&0.52&0.50&\multicolumn{1}{c}{\colorbox{\avgColor}{0.61}}&0.48\\
            &\textbf{\colorbox{\hdColor}{9}}&0.52&0.66&0.74&0.42&0.37&0.30&0.57&0.43&\rotz{-}&0.40&0.66&\multicolumn{1}{c}{\colorbox{\avgColor}{0.51}}&0.30\\
            &\textbf{10}&0.75&0.69&0.63&0.81&0.75&0.69&0.59&0.74&0.60&\rotz{-}&0.64&\multicolumn{1}{c}{\colorbox{\avgColor}{0.69}}&\textbf{0.59}\\
            &\textbf{\colorbox{\hdColor}{11}}&0.57&0.66&0.79&0.48&0.42&0.40&0.74&0.53&0.66&0.48&\rotz{-}&\multicolumn{1}{c}{\colorbox{\avgColor}{0.57}}&0.40\\[0.0cm]\cmidrule{3-13}
            &\textit{Avg}&\rotz{\colorbox{\avgColor}{0.68}}&\rotz{\colorbox{\avgColor}{0.61}}&\rotz{\colorbox{\avgColor}{0.64}}&\rotz{\colorbox{\avgColor}{0.63}}&\rotz{\colorbox{\avgColor}{0.59}}&\rotz{\colorbox{\avgColor}{0.51}}&\rotz{\colorbox{\avgColor}{0.59}}&\rotz{\colorbox{\avgColor}{0.64}}&\rotz{\colorbox{\avgColor}{0.56}}&\rotz{\colorbox{\avgColor}{0.50}}&\rotz{\colorbox{\avgColor}{0.58}}\\
            &\textit{Max}&\rotz{\textbf{0.83}}&\rotz{0.78}&\rotz{0.80}&\rotz{0.81}&\rotz{0.75}&\rotz{0.70}&\rotz{0.74}&\rotz{0.80}&\rotz{0.76}&\rotz{0.68}&\rotz{0.76}
        \end{tabular}
    }
    \caption{One-to-one transfer experiments without DA. The results show the classification accuracy. Good performance is indicated by bright green color, while bad performance by bright red color. Left-hand domains are marked in blue.}
    \label{tab:noDA}
    \end{table}
    
    All investigated transfers are in line with our previous multi-source experiment findings. The weak performance on domain 6 and 10 as target domain are reflected here as well, with lowest average classification accuracy.
    On the other hand, when training on 10, the performance on the target domains shows the best average and minimum performance, which indicates that the domain covers already a large variety in background related factor values.
    The highest max performance and also highest average performance is achieved on target domain 1. At first sight this might mostly be related to the simple white background. However, as we will show in the upcoming investigation of error causing factors, this is not the only reason.

    The overall matrix shows reoccurring patterns of weak performance. This is especially pronounced at the transfers from domain 4, 5, 6, and 8 to the domains 2, 3, 7, 9, 11. A closer examination of the these domains reveals that the objects presented there are exclusively held in either the right or the left hand. A transfer with a change in hands clearly leads to a performance drop in comparison to a transfer between the same hands. 
    Besides the hand there might be other less prominent error factors. These we like to reveal in the following.
    
\subsection{Identification of error causing factors with PCA}

\begin{figure*}[tb]
        \centering
        \input{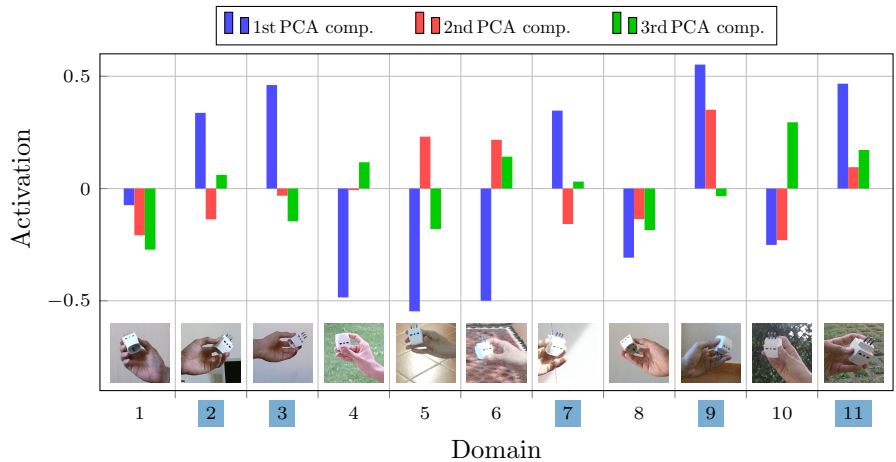}
        \caption{PCA analysis of Table\,\ref{tab:noDA}: The sign of the 1st component activation is in line with the hand that was used. Positive values refer to left hand domains (marked blue), negative values to right hand domains. Note the low absolute value for domain 1 which indicates the hand showing the object in a neutral position from the bottom. Positive values in the 2nd component could be related to domains with low contrast, while negative ones to high contrast. The 3rd component might indicate highly textured background by high positive values.}
        \label{fig:PCA}
    \end{figure*}

    In this section we try to determine error factors in a systematic way. For this we apply PCA on the one-to-one results of Table\,\ref{tab:noDA}, where each row is treated as a sample with 11 dimensions, and the gaps for transfers between the same domain are filled with 1.0 . The total variance in the accuracy is 0.26, where the first component describes already 73\% of it. The activation of the first component in Figure\,\ref{fig:PCA} shows a clear grouping in domains with high positive and high negative values. These two groups perfectly represent the used hand. Additionally, the absolute values of the activation seem to carry information how clear on one side of the target object the hand is positioned within the image. Domain 1 describes here a special case with an activation close to zero, indicating that the hand is visible mostly at a neutral position below the target object.
    
    The second component explains 15\% of the variance and shows a rather continuous distribution of activation in Figure\,\ref{fig:PCA}. We think that is related to light source factors, resulting in different average contrasts for each domain. Positive values indicate here a comparable low contrast of the objects and the holding hand to the background, which is pronounced at domain 5, 6, 9 and partly 11. 
    
    The third component represents 12\% of the variance. The  activation in Figure\,\ref{fig:PCA} is rather continuous again and might correlate with the strength of background texture, having the highest values for domains 6, 10 and 11.

    The activation of PCA components gives hints about possible domain factors. Instead of our manual interpretation, it would also be possible to correlate the PCA activation with image metadata like, time of the day, the time of the year or the GPS location of the recording. However, interpretability is not necessarily given, since the PCA suggested factorization could also be a mix of multiple human interpretable factors.   
    Besides PCA also other factorization methods like independent component analysis could have been applied, possibly leading to less clear results due to the missing orthogonality of the components.

    Note, since starting from a pre-defined domain separation on which the transfer matrix is based, PCA only allows to detect error causing factors that are consistent throughout the pre-defined domains. In other words, factors that vary within a pre-defined domain, e.\,g. on category level, can not be identified here.

\subsection{Single-Source Transfer with Domain Adaptation}
    For the transfer from one source domain to one target domain with DA, we need to additionally provide unsupervised data of the target domain. Adversarial DA will try to remove domain factors between these two domains. Before, we identified the hand as one such factor and also showed that a changing hand is the main cause of classification error. Consequently, the removal of that factor should lead to an increased performance. 

        \def\avgColor{gray!30}
    \def\hdColor{Paired-B!60}
    \renewcommand*{\MinNumber}{-0.3}%
    \renewcommand*{\MidNumber}{-0.001} %
    \renewcommand*{\MaxNumber}{0.3}%
    
    \renewcommand{\ApplyGradient}[1]{%
        \ifdim #1 pt > \MidNumber pt
            \colorbox{green!70!black}{\hphantom{-}#1}
        \else
            \colorbox{red!80!white}{#1}
        \fi
    }
    
    \setlength{\fboxsep}{1mm} 
    \newcolumntype{U}{>{\collectcell\ApplyGradient}c<{\endcollectcell}}

    \setlength{\tabcolsep}{0.2pt} 
    \renewcommand{\arraystretch}{1.1} 
    \begin{table}[tbp]
        \centering
        \resizebox{\columnwidth}{!}{
        \begin{tabular}{cc*{11}{U}c@{\hskip 0.07in}c}
            \multicolumn{2}{c}{} & \multicolumn{11}{c}{\textbf{Target Domain} (w/ usv data for training)} \vspace{0.1cm} \\ 
            &\multicolumn{1}{c}{}& \rotz{\textbf{1}} &
             \rotz{\textbf{\colorbox{\hdColor}{2}}} &
             \rotz{\textbf{\colorbox{\hdColor}{3}}} &
             \rotz{\textbf{4}} &
             \rotz{\textbf{5}} &
             \rotz{\textbf{6}} &
             \rotz{\textbf{\colorbox{\hdColor}{7}}} &
             \rotz{\textbf{8}} &
             \rotz{\textbf{\colorbox{\hdColor}{9}}} &
             \rotz{\textbf{10}}& 
             \rotz{\textbf{\colorbox{\hdColor}{11}}}&
             \hphantom{a}\textit{Avg}\hphantom{a}&
             \textit{Min}\\ \cmidrule{3-13}
            \parbox[t]{4mm}{\multirow{11}{*}{\rotatebox[origin=tr]{90}{\textbf{Source Domain}}}}
            &\textbf{1}&\rotz{-}&0.08&-0.06&0.04&0.07&0.21&-0.01&0.04&-0.07&0.22&0.09 &\multicolumn{1}{c}{ \colorbox{\avgColor}{\hphantom{-}0.06}}&\textbf{0.49}\\
            &\textbf{\colorbox{\hdColor}{2}}&0.04&\rotz{-}&0.02&-0.02&-0.02&-0.07&0.04&-0.02&-0.04&0.05&0.03 &\multicolumn{1}{c}{ \colorbox{\avgColor}{\hphantom{-}0.00}}&0.48\\
            &\textbf{\colorbox{\hdColor}{3}}&-0.03&0.02&\rotz{-}&-0.11&-0.17&-0.02&0.02&-0.10&0.01&0.06&0.07 & \multicolumn{1}{c}{\colorbox{\avgColor}{-0.02}}&0.33\\
            &\textbf{4}&0.05&0.00&-0.05&\rotz{-}&0.12&0.06&-0.05&0.07&-0.08&0.05&-0.03 & \multicolumn{1}{c}{\colorbox{\avgColor}{\hphantom{-}0.01}}&0.28\\
            &\textbf{5}&0.04&0.06&-0.05&0.02&\rotz{-}&0.09&-0.03&0.05&-0.05&0.12&0.00 & \multicolumn{1}{c}{\colorbox{\avgColor}{\hphantom{-}0.02}}&0.32\\
            &\textbf{6}&0.08&0.06&-0.03&0.10&0.14&\rotz{-}&-0.01&0.11&-0.08&0.17&-0.02 & \multicolumn{1}{c}{\colorbox{\avgColor}{\hphantom{-}0.05}}&0.31\\
            &\textbf{\colorbox{\hdColor}{7}}&-0.09&0.04&0.02&-0.09&-0.09&0.02&\rotz{-}&-0.16&-0.01&0.03&0.02 & \multicolumn{1}{c}{\colorbox{\avgColor}{-0.03}}&0.43\\
            &\textbf{8}&0.00&-0.04&-0.08&0.03&0.09&0.08&-0.07&\rotz{-}&-0.08&0.17&-0.04 & \multicolumn{1}{c}{\colorbox{\avgColor}{\hphantom{-}0.01}}&0.41\\
            &\textbf{\colorbox{\hdColor}{9}}&0.05&0.10&0.07&-0.08&-0.05&-0.02&0.11&-0.02&\rotz{-}&-0.02&0.11 & \multicolumn{1}{c}{\colorbox{\avgColor}{\hphantom{-}0.03}}&0.28\\
            &\textbf{10}&0.03&-0.05&-0.13&0.01&0.00&0.01&-0.07&0.01&-0.15&\rotz{-}&-0.05 & \multicolumn{1}{c}{\colorbox{\avgColor}{-0.04}}&0.45\\
            &\textbf{\colorbox{\hdColor}{11}}&-0.04&0.01&-0.01&0.00&-0.09&-0.03&0.00&-0.11&0.07&-0.02&\multicolumn{1}{c}- &\multicolumn{1}{c}{ \colorbox{\avgColor}{-0.02}}&0.34\\ \cmidrule{3-13}
            & \textit{Avg} & \rotz{\colorbox{\avgColor}{\hphantom{-}0.01}} & \rotz{\colorbox{\avgColor}{\hphantom{-}0.03}} & \rotz{\colorbox{\avgColor}{-0.03}} & \rotz{\colorbox{\avgColor}{-0.01}} & \rotz{\colorbox{\avgColor}{\hphantom{-}0.00}} & \rotz{\colorbox{\avgColor}{\hphantom{-}0.03}} & \rotz{\colorbox{\avgColor}{-0.01}} & \rotz{\colorbox{\avgColor}{-0.01}} & \rotz{\colorbox{\avgColor}{-0.05}} & \rotz{\colorbox{\avgColor}{\hphantom{-}0.08}} & \rotz{\colorbox{\avgColor}{\hphantom{-}0.02}} & \rotz{}\\
            & \textit{Max} &\rotz{\hphantom{-}0.83}&\rotz{\hphantom{-}0.80}&\rotz{\hphantom{-}0.82}&\rotz{\hphantom{-}0.83}&\rotz{\hphantom{-}0.83}&\rotz{\hphantom{-}0.76}&\rotz{\hphantom{-}0.75}&\rotz{\textbf{\hphantom{-}0.84}}&\rotz{\hphantom{-}0.75}&\rotz{0.74}&\rotz{\hphantom{-}0.78}
        \end{tabular}
    }
    \caption{One-to-one transfer experiments with DA. The difference in percentage points to Table\,\ref{tab:noDA} is shown. \textit{Min} and \textit{Max} refer to absolute accuracies. Left-hand domains are marked in blue.}
    \label{tab:DAvsnoDA}
    \end{table}
    
    However, when looking at the results in Table\,\ref{tab:DAvsnoDA}, usually a further decrease in performance is visible for changing hands.
    This is especially the case for the transfer from domain 3 to 5 with a decrease of 17 percentage points and 7 to 8 with 16 percentage points.    
    An increase is mainly observed on transfers where the hand does not change between domains. The biggest improvements can be found in the transfers from domain 1 to 10 and 1 to 6 with an increase of 22 and 21 percentage points. We think that this significant increase is also caused by the same effect as in the multi-source experiments, where the unsupervised data of 6 and 10 has provided the biggest benefit by introducing the new factor values related to the background.
    %
    The average increase in accuracy for transfers where the hand does not change is at 5.7 percentage points, while the average decrease for changing hands is -3.8 percentage points.
    
    The highest minimum performance has dropped significantly from former 59\% accuracy on source domain 10 to now 49\% on domain 1, indicating the strong negative transfer caused by DA. 
    
    Our results clearly indicate strong negative transfer, caused by removing the factor \textit{'hand'} through DA. The reason for this is on one side that the hand is simultaneously a task-informative factor, by its posture giving clues about the object category. On the other side it has visual closeness to the target object, which means removing it can harm classification significantly due to fixed receptive field sizes in the architecture.
    
\subsection{FP-DA for Multi-Source Domain Adaptation}
    \begin{figure*}[tb]
        \centering
        \def\plotheight{0.5\columnwidth} 
\def\plotwidth{1\columnwidth} 
\def\legendscale{0.7}
\def\plotTextSize{}
\def\drawParam{none}
\pgfplotsset{every tick label/.append style={font=\footnotesize}} 
\begin{tikzpicture}[font=\plotTextSize]
		\begin{axis}[
        ybar=0pt,
        bar width=2pt,
		legend style={nodes={scale=\legendscale, transform shape}},
		height=\plotheight,
		width=\plotwidth,
		grid=minor,
		ymin = 70,
		ymax = 100,
		xtick={1,2, 3, 4, 5, 6, 7, 8, 9, 10, 11, 12, 13, 14},
        xticklabels={1,2, 3, 4, 5, 6, 7, 8, 9, 10, 11, , \footnotesize \textit{Avg},  \footnotesize \textit{Min}},
        minor xtick={1.5, 2.5, 3.5, 4.5, 5.5, 6.5, 7.5, 8.5, 9.5, 10.5, 11.5, 12.5, 13.5},
        minor ytick={80, 90},
        major tick length=0pt,
        xtick pos =left,
        ytick pos = left,
		xlabel style={text width=9cm, align=center},
		xlabel= Target Domain,
		ylabel= Classification Accuracy,
		ylabel style = {yshift=-0.4cm},
		legend pos=south west,
		legend cell align={left},
        legend columns=3,
        legend style={/tikz/every even column/.append style={column sep=0.25cm}},
        legend style={at={(0.5,1.025)},anchor=south},
        enlarge x limits = {abs=15pt},
		yticklabel=\pgfmathparse{\tick/100}\pgfmathprintnumber{\pgfmathresult}\,,
		]
		\centering
		\addplot[fill=Paired-B, draw = \drawParam] coordinates {
            (1, 94.32)
            (2, 91.30)
            (3, 94.9)
            (4, 93.0)
            (5, 90.6)
			(6, 75.3)
			(7, 89.02)
			(8, 94.40)
			(9, 90.8)
            (10, 81.1)
            (11, 92.9)
            (13, 89.78) 
            (14, 75.3) 
		};
		\addlegendentry{no DA}
        
        \addplot[fill=black, draw = \drawParam] coordinates {
            (1, 92.6)
            (2, 91.8)
            (3, 92.9)
            (4, 92.3)
            (5, 91.8)
            (6, 76.2)
            (7, 87.54)
            (8, 94.0)
            (9, 89.7)
            (10, 80.2)
            (11, 91.9)
            (13, 89.17) 
            (14, 76.2)
        };
        \addlegendentry{DA\,\cite{Ganin15}\,(w/o usv)}
        
        \addplot[fill=Paired-C, draw = \drawParam] coordinates {
            (1, 92.1)
            (2, 92.9)
            (3, 92.6)
            (4, 90.5)
            (5, 91.9)
            (6, 86.34)
            (7, 87.27)
            (8, 92.39)
            (9, 85.7)
            (10, 87.03)
            (11, 89.91)
            (13, 89.87) 
            (14, 85.7) 
        };
        \addlegendentry{DA\,\cite{Ganin15}\,(w/ usv)}

		\addplot[fill=Paired-D, draw = \drawParam] coordinates {
            (1, 93.7)
            (2, 95.14)
            (3, 94.5)
            (4, 94.15)
            (5, 92.99)
            (6, 86.8)
            (7, 89.51)
            (8, 94.38)
            (9, 90.07)
            (10, 87.1)
            (11, 91.8)
            (13, 91.83) 
            (14, 86.8) 
        };
        \addlegendentry{FP-DA\,(1st, w/ usv)}
        
        \addplot[fill=Paired-E, draw = \drawParam] coordinates {
            (1, 93.2)
            (2, 94.8)
            (3, 93.8)
            (4, 93.4)
            (5, 92.1)
            (6, 85.8)
            (7, 88.3)
            (8, 94.0)
            (9, 89.4)
            (10, 87.0)
            (11, 91.0)
            (13, 91.16) 
            (14, 85.8) 
        };
        \addlegendentry{FP-DA\,(2nd, w/ usv)}

        \addplot[fill=Paired-F, draw = \drawParam] coordinates {
            (1, 92.8)
            (2, 95.0)
            (3, 93.4)
            (4, 93.1)
            (5, 92.2)
            (6, 85.6)
            (7, 88.5)
            (8, 93.8)
            (9, 89.4)
            (10, 87.1)
            (11, 91.3)
            (13, 91.09) 
            (14, 85.6) 
        };
        \addlegendentry{FP-DA\,(3rd, w/ usv)}

        \addplot[fill=Paired-G, draw = \drawParam] coordinates {
            (1, 92.89)
            (2, 95.13)
            (3, 93.5)
            (4, 93.46)
            (5, 92.23)
            (6, 85.6)
            (7, 89.0)
            (8, 93.6)
            (9, 89.08)
            (10, 86.74)
            (11, 91.37)
            (13, 91.14) 
            (14, 85.6) 
        };
        \addlegendentry{FP-DA\,(rand., w/ usv)}

 		\addplot[fill=Paired-H, draw = \drawParam] coordinates {
            (1, 93.43)
            (2, 92.23)
            (3, 94.86)
            (4, 92.95)
            (5, 90.79)
            (6, 75.74)
            (7, 89.1)
            (8, 93.99)
            (9, 89.97)
            (10, 81.4)
            (11, 91.29)
            (13, 89.61) 
            (14, 75.74) 
        };
        \addlegendentry{FP-DA\,(1st, w/o usv)}

		\end{axis}
\end{tikzpicture}
        \caption{Multi-source DA with our proposed FP-DA. 1st, 2nd, and 3rd describe the grouping based on the three PCA components, rand.\ a random permutation. FP-DA with the domain grouping based on the used hand (1st) and usv data shows the highest average and highest minimum performance among all constellations. 
        }
        \label{fig:leave_one_out}
    \end{figure*}
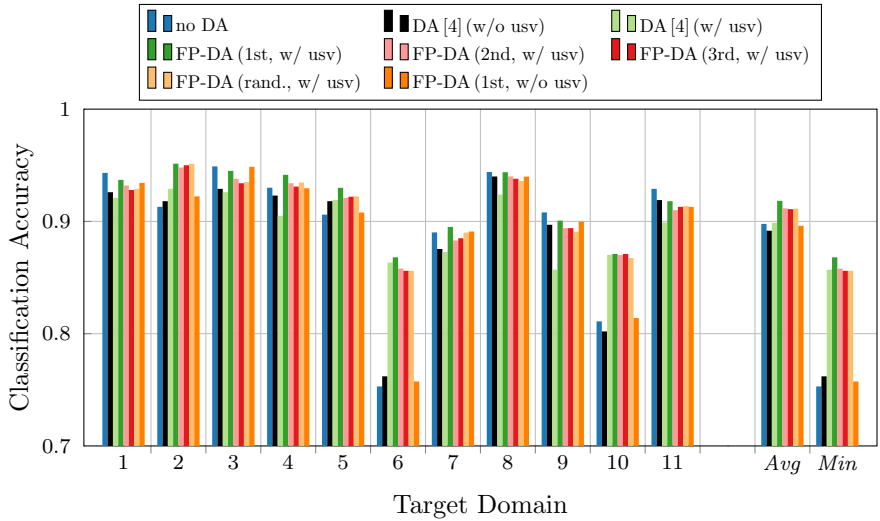
    %
    %
    In this section we like to investigate how FP-DA can reduce negative transfer by preserving a given factor. The best candidate for such a factor is the hand, which also corresponds to the first PCA component. Similarly, we will also evaluate the groupings defined by the sign of the activation of the second and third PCA eigenvector. As FP-DA switches off competition between certain pairs of domains, in general it leads to a decreased influence of the domain classification path. To investigate this effect, we also tested several random factors, each having 2 groups with 5 respectively 6 domains. The results, together with the previous multi-source experiments, can be found in Figure\,\ref{fig:leave_one_out}.
    
    Looking at the performance when defining the groups based on the 1st PCA component, i.\,e. based on the hand used for presenting, the overall highest average and highest minimum accuracy can be reached. The increase in comparison to DA\,\cite{Ganin15} with unsupervised data is at 1.9 and 1.1 percentage points, respectively. The detailed results for each domain reveal that FP-DA manages mostly to reduce the negative transfer that was caused by standard DA, in certain cases an even better classification accuracy than without DA could be achieved. This applies for example to domain 4 and 7. Overall, here the expectations of FP-DA were fulfilled and its effectiveness was proven.
    
    Using the 2nd or the 3rd PCA component provides only a reduced gain of average performance of about 1.2 percentage points. A similar gain is achieved for the random groupings. This indicates that in general a reduced competition and thus influence of the domain path is beneficial here, but not the removal of 2nd and 3rd PCA component in particular.
    
    Another strategy to deal with the changing hand would be to augment the data by horizontal flipping. However, this is not a method to deal with domain factors in general and was therefore not considered here.
\section{Conclusion}

    In this work we discussed that current domain transfer literature lacks of more specific investigations of causes of negative transfer. Using only different data distributions as explanation for effects in DA is in our opinion not sufficient.
    For in-depth analysis we introduced in this paper our factor theory, which explains the appearance of an image by a mixture of multiple factors with different values that describe a scene.
    We discussed that in general there is no unique factorization of a given dataset, however, a well-chosen factorization can help to describe characteristics in more detail. With focus on DA datasets, certain factors can be categorized as domain-informative and domains can be described by a combination of these factors, the domain factors.
    For a given classification task, certain factors can further be described as task-informative. With these categorizations we used the factor theory to explain effects of machine learning and DA, where the challenge lies mainly in new factor values or new combinations of such in the test data.
    
    The standard adversarial DA approach\,\cite{Ganin15}, targets to remove all domain factors from the feature representation to become domain invariant. We discussed that this can lead to negative transfer if such factors are not only domain-informative, but also task-informative, or visually close to task-informative factors.
    We showed this in extensive experiments, where we first saw decreased performances in certain multi-source constellations caused by DA and later investigated this in detail by one-to-one experiments. There we found, supported by a PCA analysis, that without DA the hand holding the object was the most error causing factor in the transfer between domains where different hands were used. The subsequent application of DA on these transfers showed a further decrease in performance by the removal of this domain factor \textit{'hand'} and confirmed our hypothesis about negative transfer caused by DA. We think that the removal of this domain factor also removed the task-informative factor \textit{'hand posture'}.
    
    To actively preserve such factors and consequently reduce negative transfer, we proposed FP-DA, a training method for multi-source DA that allows only competition within groups of domains that have the same value for the factor to preserve.
    Using FP-DA, we showed in our experiments that by preserving the factor \textit{'hand'} the negative transfer could be reduced and the highest average and highest minimum classification accuracy could be achieved.


\bibliography{bibtex_file}

\end{document}